\pdfoutput=1

\documentclass[11pt]{article}

\usepackage[]{acl}

\usepackage{times}
\usepackage{latexsym}

\usepackage[T1]{fontenc}

\usepackage[utf8]{inputenc}

\usepackage{microtype}

\usepackage{float}
\usepackage{hyperref}

\title{PoolNet: Deep Learning for 2D to 3D Video Process Validation}

\author{Sanchit Kaul \\
  University of California, \\
  Davis \\
  \texttt{skkaul@ucdavis.edu} \\\And
  Joseph Luna \\
  University of California, \\
  Davis \\
  \texttt{jmiluna@ucdavis.edu} \\\And
  Shray Arora \\
  University of California, \\
  Davis \\
  \texttt{rayarora@ucdavis.edu}\\}

\begin{document}
\maketitle
\begin{abstract}
Lifting Structure-from-Motion (SfM) information from sequential and non-sequential image data is a time-consuming and computationally expensive task. In addition to this, the majority of publicly available data is unfit for processing due to inadequate camera pose variation, obscuring scene elements, and noisy data. To solve this problem, we introduce PoolNet, a versatile deep learning framework for frame-level and scene-level validation of in-the-wild data. We demonstrate that our model successfully differentiates SfM ready scenes from those unfit for processing while significantly undercutting the amount of time state of the art algorithms take to obtain structure-from-motion data. All code related to this paper can be found at \href{https://github.com/sanchitkaul/PoolNet.git}{https://github.com/sanchitkaul/PoolNet.git}
\end{abstract}

\section{Introduction}

Structure-from-Motion (SfM) pipelines like COLMAP \cite{schoenberger2016mvs}, and GLOMAP \cite{pan2024glomap} have helped make scalable 3D reconstruction widely accessible. Their application spans across domains such as robotics, AR/VR, autonomous navigation, and virtual scene synthesis; however, COLMAP tends to be highly sensitive to the image quality of the input, overlap between images, and diversity in viewpoints. 

While tools like photogrammetry, Gaussian splatting \cite{kerbl3Dgaussians}, and Neural Radiance Fields (NeRF) \cite{mildenhall2020nerf} rely on accurate camera pose and image depth information, large scale datasets scraped from public sources often lack metadata, increasing computational cost and noise. In practice, when running the pipeline especially against web scraped sources, it often leads to hours of wasted computation per failed run and thereby, failed reconstructions.

Our project, PoolNet, aims to mitigate this problem by building a model that can predict, ahead of time, whether a sequence of images is likely to result in a successful reconstruction. By avoiding running entire SfM pipelines on data that would likely not provide usable 3D results, it would improve scalability and resource usage in big data environments. Through the analysis of metrics such as frame usage fraction, point cloud size and camera pose diversity, our model will determine if passing the data through COLMAP, or other similar structure from motion (SfM) tools, is worthwhile. This way, we aim to reduce wasted GPU cycles and prioritize valuable data in attempts to scrape 3D elements from public video data, or image collections. 

Poolnet’s filtered output has a wide-ranging real world impact. For instance, Gaussian Splatting, a real-time neural rendering technique for point-based 3D scenes, benefits from pre-filtered datasets that reduce reconstruction noise and improve visual clarity and geometric accuracy. More generally, any pipeline that utilizes sparse geometry can benefit from such a filter that saves compute and prioritizes promising inputs. 

In order to effectively estimate reconstruction success, we pose the problem as a supervised learning task over sequences of frames. Each scene is labeled with outcome metrics from a COLMAP run, or a photogrammetry mesh. Our architecture operates on per-frame image features, aggregates them and outputs predictions that correlate with SfM success. This approach allows us to selectively pass only high quality sequences to downstream reconstruction pipelines, and discard noisy inputs early on. 

The rest of this paper outlines our design in detail. We begin by defining our problem setup and learning objectives, followed by an explanation of our dataset collection and label generation pipeline. We then present our model architecture, and finally evaluate the effectiveness of PoolNet’s predictions through experimental results.

\section{Method}

\subsection{Problem definition}
As mentioned earlier, COLMAP requires significant compute and is sensitive to variations in image quality, camera poses, and scene overlap. For instance, a failed COLMAP run of a 1{,}353-image dataset on an RTX 3080 Ti with 12GB VRAM may stall during the reconstruction process and run for approximately 8+ hours before timing out~\cite{pan2024glomap}.

Our goal is to predict, before running the SfM pipeline, whether a scene is likely to result in a usable reconstruction. This problem can be framed as a supervised learning task: given a sequence of frames 
\[
F = (f_0, f_1, \ldots, f_n),
\]
we aim to predict targets associated with COLMAP’s output. These include:

\begin{itemize}
    \item $T_s$: A binary label indicating whether COLMAP produces a valid sparse reconstruction.
    \item $T_o$: The fraction of input frames successfully used in the output model (i.e., the number of images used in the final model divided by the total input count).
    \item $V_t$, $V_r$: Pose diversity metrics, measuring translational and rotational variability between recovered camera poses.
\end{itemize}

These targets allow us to model reconstruction quality. $T_s$ serves as the primary classification label, while $V = \{V_t, V_r\}$ provides secondary regression targets that encode geometric variation in successful reconstructions.

One of the key challenges is that most input sequences partially succeed, and complete failures are rare. To mitigate this class imbalance, we apply a threshold to $T_o$ to more strictly define failure, and incorporate additional geometric cues to strengthen the predictive signal.

\subsection{Datasets}
For the data, we used a large dataset of object scans \cite{choi2016redwood}, an RGB-D dataset which includes photogrametric reconstructions of the objects, and as well as object categories. Because the dataset was crowdsourced from public users without experience in 3D scanning, the dataset features a wide variety of image quality. It includes videos with differing lighting conditions, lens flaring, motion blur, and varying viewpoints. Additionally, the modality of the data is consistent and compact, making it a good choice for real world SfM challenges.

\subsection{Label Generation via COLMAP}
To generate the labels for our training, we ran COLMAP’s sparse reconstruction pipeline on each video sequence. However, COLMAP often generates multiple sparse reconstructions per scene. Initially, we chose one of these reconstructions randomly. However, this introduced high variance and generated noisy labels that our model struggled to learn from. To fix this, we modified our label generation pipeline to pick the sparse model with the highest number of reconstructed images, and only use metrics derived from that. This decision improved consistency and provided a much stronger signal.

To further improve the label quality, we redefined what we counted as a successful reconstruction. Instead of treating any COLMAP output as success, we only considered a scene successful if at least 45 percent of the input images were used in the final sparse model. This further resulted in a more balanced classification signal.

Additionally, we noticed training directly on the geometric metrics did not provide good results. To counteract this, we utilized cosine embedding loss which helps in pulling the embeddings of images from the same scene closer while pushing apart unrelated ones. This approach helped stabilize learning and boosted final performance.

Lastly, to enable fair comparisons of our model’s inference time against other systems that perform 3D reconstruction, we aligned our input sequence lengths with those used in the ETH3D benchmark, specifically the ones evaluated in the Glomap paper. Matching their sequence lengths allowed us to directly benchmark runtime and see how our approach stacks up.

\subsection{Model Architecture}
Our model architecture went through multiple iterations before we landed on the final setup. Initially, we implemented a pre-trained ViT (Vision Transformer) \cite{wu2020visual} as the image encoder along with a LSTM prediction head to capture dependencies among sequences. The overarching idea involved utilizing the pretrained ViT to encode individual frames and the LSTM to aggregate spatial patterns across image sequences.

However, this setup didn’t perform well. The ViT requires a large amount of data and compute for effective fine tuning. Our dataset, which did not contain a complete set of labels, and contained only 10,933 unique scenes was not fit for fine tuning a model of the size of the vision transformer. When testing this setup, we simply utilized embeddings generated by the vision transformer without fine tuning. As a result, the LSTM was not fed a good signal, and was not able to achieve convergence. We hypothesize that the classification task that the ViT was originally trained for functioned as a poor proxy for learning temporally relevant geometric features.

After testing, we shifted to a CNN based image encoder along with a transformer predictor to model the dependencies between frames. We shifted to this architecture for two main reasons:

\begin{enumerate}
    \item CNNs are easier to train from scratch, tending to converge more quickly when working with moderately sized datasets, while still doing a good job in extracting useful visual features from images.
  
    \item Unlike LSTMs, transformers are good at modeling long term relationships. This is in alignment with the way COLMAP works as reconstruction is highly dependent on coverage and diversity rather than sequence order.
\end{enumerate}

So, the current architecture consists of:

\begin{itemize}
    \item CNN encoder: Input frames are passed through a lightweight convolutional encoder that maps it to a 64-dimensional latent vector. This replaces the earlier ViT, and has proven much more stable during training.
\end{itemize}
\begin{itemize}
    \item Transformer Sequence Model: This per frame encoded data is passed into a Transformer that learns how separate frames connect over time. It aggregates across all input sequences, learning to detect global configurations that correlate with successful or failed reconstructions.
\end{itemize}

Unlike our earlier models, this setup learns much better. The CNN provides reliable features while the Transformer can pick up patterns like poor view coverage or redundant angles. Since reconstruction failure often stems from these issues, the model is able to generalize better.
We're currently training the model using a balanced binary classification loss (success vs failure), where the labels are derived from whether the Redwood group was able to generate a mesh from a given COLMAP scene. Early results suggest an accuracy of ~80 percent on the test split, which is a significant improvement over prior iterations.

\subsection{Training Methodology}

In order to combat issues relating to vanishing and exploding gradient, we separate our training protocol into multiple separate phases. We first train our convolutional encoder model to understand scene elements. We then freeze this model and train our transformer on it's own separate binary classification task. 
During the embedder pre-training phase, we utilize both scenes that were and were not reconstructed successfully indiscriminately. At each learning iteration, our trainer follows the proceeding steps:
\begin{enumerate}
    \item Select a random image from any scene
    \item Randomly select either an image from the same scene, or from any other scene in the dataset with a 50/50 ratio
    \item Repeat steps 1 and 2 for each item in the batch
    \item Compute a forward pass on the convolutional model, encoding both images to a 64 dimensional latent space.
    \item For same-scene image pairs, we utilize Cosine Similarity Loss to encourage small cosine angles, while encouraging large angles in differen-scene pairs.
\end{enumerate}

Our theory in utilizing this experimental setup is that the model will learn to understand how unique views relate to the same scene while ignoring image from other scenes. This model alone may also be useful in real-world scraping tasks for understanding if unique images from different users may be related to eachother (eg. if two people take pictures of Times Square in new york, we may be able to utilize both images for training, despite their being from different sources.)
While training our transformer model, we simplified our training task to include only the binary success metric $T_s$. Initially, we attempted to measure the overlap between images by computing the average angle between the camera-point-camera corner of point matches, however initial testing showed that this metric provided a weak signal to our learner. We hope to revisit the concept of image overlap in the future, utilizing improved metrics and model pre-training to boost the clarity of the signal. For now however, we were successfully able to train our model to differentiate between successful and unsuccessful scenes utilizing the following setup.
\begin{enumerate}
    \item Select a random sequence of 10 images from a scene
    \item Repeat for the size of the batch
    \item Obtain image embeddings from the convolutional model
    \item Apply a transformer model to the image embeddings, utilizing learned positional embeddings
    \item Regress based on the correct classification of the scene as successfully, or unsuccessfully reconstructed

Utilizing this simplified training setup, we were able to achieve convergence, obtaining a model with 79.8\% accuracy
\end{enumerate}

\section{Experiments}

We first evaluate our convolutional embedding model on it's pre-training task to validate that the quality of its embedding space and evaluate its potential as a frame-proposal network. We also evaluate our transformer model based on its accuracy at predicting binary ability to reconstruct to validate it's effectiveness. Finally, we compare the inference time of PoolNet to state-of-the-art methods for for generating structured scenes from image collections. 

\subsection{Frame Proposal}
During training, to ensure that our model is learning an effective embedding space, we test our image pairing model's accuracy on unseen test data. We utilize a 9:1 train:test split in our image data, utilizing both scenes that could and could not be reconstructed. We found that after 1 epoch of training on the scene dataset, we were able to achieve an accuracy of 74.3\% on previously unseen data. This validated that our model was learning a generalizable embedding space, and confirmed that there were likely to be geometrically relevant information within our embedding space for our transformer model to learn from.

\subsection{Scene Proposal}
During our initial attempts to train our model, we evaluated the model based on it's output diversity, loss, and accuracy. When trying to train directly on our geometric labels, we were unable to achieve convergenc, with our model continually trending towards the mean value of the labels, even after implementing regularization to encourage diverse outputs. 
After simplifying our training task, utilizing pre-training, we had significantly more success in our training. first balanced our classes, as only 430 of the 10,933 scenes were possible to compute meshes from. We randomly select 430 "meshable" and 430 "non-meshable" scenes and combine them into a dataset of 860 scenes before performing a 9:1 train:test split. After training, we test our model using our test data to find the previously mentioned 79.8\% accuracy.
\subsection{Time Savings}
In order to evaluate the effectiveness of our model at reducing compute time in real-world data scraping scenarios, we compare the inference time of our model to the completion time of numerous structure-from-motion tools \cite{pan2024glomap}. In order to make a fair comparison to these metrics, we test our compute time over the entire image dataset, rather than utilizing subsetting of each scene, as was done during training. Even with this significant reduction to speed, we show that on average, our model takes only 17.0\% as long as the next fastest method, Theia \cite{theia-manual}. We show that even under adversarial testing conditions, our model is capable of significantly undercutting the amount of compute time needed to validate in-the-wild data.
\noindent
\begin{table}[h]
\footnotesize  
\setlength{\tabcolsep}{1.5pt}  
\begin{tabular}{lccccc}
\hline
\textbf{Dataset} & \textbf{OpenMVG} & \textbf{Theia} & \textbf{GLOMAP} & \textbf{COLMAP} & \textbf{Ours} \\
\hline
Cables & 103.1 & 339.5 & 195.6 & 2.553e3 & \textbf{13.0} \\
Camera & \textbf{1.5} & 5.3 & 10.0 & 196.2 & 3.9 \\
Ceiling & 78.3 & 52.6 & 111.0 & 1.058e3 & \textbf{19.2} \\
Desk & 376.2 & 195.3 & 150.0 & 1.115e3 & \textbf{25.9} \\
Einstein & 150.3 & 70.5 & 142.1 & 1.231e3 & \textbf{6.1} \\
Kidnap & 114.4 & 356.7 & 144.3 & 731.2 & \textbf{11.3} \\
Large & 91.9 & 60.2 & 77.6 & 983.8 & \textbf{17.8} \\
Mannequin & 33.5 & 29.7 & 44.3 & 301.2 & \textbf{8.1} \\
Motion & 8.597e2 & 109.0 & 7.889e2 & 9.995e3 & \textbf{30.6} \\
Planar & 313.8 & 167.5 & 533.3 & 2.350e3 & \textbf{7.9} \\
Plant & 21.7 & 35.7 & 28.7 & 202.7 & \textbf{1.0} \\
Reflective & 7.213e2 & 118.3 & 4.344e2 & 6.574e3 & \textbf{55.8} \\
Repetitive & 63.2 & 136.8 & 74.5 & 561.1 & \textbf{25.0} \\
SFM & 91.7 & 170.5 & 239.7 & 469.6 & \textbf{8.1} \\
Sofa & 9.1 & \textbf{8.9} & 10.1 & 157.3 & 12.2 \\
Table & 182.0 & 97.8 & 221.5 & 2.778e3 & \textbf{14.1} \\
Vicon & 50.9 & 88.2 & 46.2 & 474.8 & \textbf{6.2} \\
\hline
\textbf{Average} & 120.8 & 91.8 & 133.5 & 1.115e3 & \textbf{15.6} \\
\hline
\end{tabular}
\caption{Time Savings (in seconds). Large values are shown in scientific notation for brevity.}
\label{tab:time_savings}
\end{table}

\section{Related Work}

\subsection{Structure from Motion and 3D Reconstruction}
Structure from Motion (SfM) is a widely-used pipeline for recovering 3D structure from 2D images by estimating camera poses and reconstructing 3D structure from matched feature points across multiple images. COLMAP \cite{schoenberger2016sfm} is a leading open-source SfM system, known for its robustness and extensibility. Recent large-scale SfM approaches like Glomap \cite{pan2024glomap} improve reconstruction quality using global feature graphs and better initialization. However, these methods still rely on  well-structured datasets, making them brittle in real-world scraping scenarios.
\subsection{Image and Video Encoding Models}
Previous approaches to encoding unstructured visual data in SfM pipelines have utilized Vision Transformers (ViTs) \cite{wu2020visual} and LSTM modules \cite{zhou2023continual} to model spatial and temporal patterns, respectively. ViTs, which treat images as sequences of fixed-size patches, offer scale-invariant encodings that generalize well across diverse input conditions. Meanwhile, LSTMs have historically been favored for modeling sequential dependencies in video frames due to their ability to retain long-range temporal context.
However, for our task predicting SfM viability on variable and low-quality video data , these methods did not signal well in practice. Instead, we adopt a CNN + Transformer architecture.. The CNN works as a lightweight frame feature extractor, while the Transformer models dependencies across the sequence. This design enables our model to capture both local geometric details and long range context, resulting in stronger performance on noisy data.

\subsection{Filtered Dataset Utility and Downstream Use}
Good quality 3D reconstructions are foundational to photogrammetry and robotics. Methods like NeRF \cite{mildenhall2020nerf} and Gaussian Splatting \cite{kerbl3Dgaussians} rely on accurate 3D scene geometry. However, both are sensitive to poor-quality reconstructions and noisy camera poses. PoolNet aims to filter out such sequences before intensive pipelines with large compute are applied, thereby saving time, reducing resource waste, and improving downstream quality.
\subsection{Data Filtration in Large-Scale Crawls}
Most real-world 3D datasets (e.g., ScanNet, ETH3D, MegaDepth) are built through extensive post-hoc cleaning. These steps are often hard to reproduce when scaling to large video crawls where many sequences lack sufficient viewpoint overlap or consistent quality. Some works propose post-reconstruction filtering \cite{lindenberger2021pixelperfect}, but few attempt to predict the feasibility of reconstruction \textit{prior} to running SfM. Our work bridges this gap by creating a filtration model based on COLMAP outputs. This pre-filtering approach reduces failed reconstruction attempts and enables the scalable use of uncurated, internet-scale video collections.

\section{Conclusion}

In this paper, we present the problem of validating large quantities of in-the-wild image and video data. To solve this problem, we present PoolNet, an architecture for frame matching and scene validation with regards to 3D camera pose estimation and reconstruction. Our methodology utilizes a contrastive approach for pairing images from the same 3D environment. This model may be particularly useful for reconstructing models of commonly photographed scenes, such as tourist attractions, and may be especially useful for those attempting to digitally archive objects of historical value. In addition to our convolutional image matcher, we utilize a transformer model for extraction of geometric features from our first model's embeddings, and determining whether or not a scene will be able to be reconstructed from an arbitrary set of images. We believe that this completed model will be of use for those trying to create large 3D datasets from publicly available sources. 

While we were able to achieve noteworthy results utilizing our model, we feel that there is significant room for iteration upon our current framework. First, we would like to address the issue of geometric image comparison. Our initial attempts to utilize statistics relating to the epipolar geometry of images have shown little success. Despite this, we believe that with improved pre-training and larger datasets, we may be able to utilize transformer-based architecture to approximate geometric properties of images in 3D space. Second, we would like to spend time expanding our dataset. While we were able to run Colmap on ~408 scenes from the redwood dataset, we would like to both expand the total number of scenes run, and increase percentage of scenes that achieved effective reconstruction. This would allow us to increase the amount of data we can represent in our training, improve the validation of our results, and potentially use deeper, more data-hungry models such as transformers for the encoding backend of our architecture. Finally, we would like to spend time adjusting the training parameters of our model. Due to time restraints, we were forced to reduce the training time of our final model set to around 8 hours. We believe that by increasing the training time of our model, altering the parameter count and depth of both our transformer and convolutional models, increasing our dataset size, and tuning our hyperparameters, we could significantly increase our model's performance. 

Beyond improving the performance of our existing models, we hope to utilize this architecture in future efforts to create large, diverse 3D scene datasets from public data. Construction of a dataset such as this could have far-reaching applications in fields like 3D generative AI, 3D scene understanding, temporal image modeling, and 3D Language Embedding.

\bibliography{anthology,custom}

@inproceedings{schoenberger2016sfm,
    author={Sch\"{o}nberger, Johannes Lutz and Frahm, Jan-Michael},
    title={Structure-from-Motion Revisited},
    booktitle={Conference on Computer Vision and Pattern Recognition (CVPR)},
    year={2016},
    url={https://demuc.de/papers/schoenberger2016sfm.pdf}
}

@inproceedings{schoenberger2016mvs,
    author={Sch\"{o}nberger, Johannes Lutz and Zheng, Enliang and Pollefeys, Marc and Frahm, Jan-Michael},
    title={Pixelwise View Selection for Unstructured Multi-View Stereo},
    booktitle={European Conference on Computer Vision (ECCV)},
    year={2016},
}

@inproceedings{pan2024glomap,
    author={Pan, Linfei and Barath, Daniel and Pollefeys, Marc and Sch\"{o}nberger, Johannes Lutz},
    title={{Global Structure-from-Motion Revisited}},
    booktitle={European Conference on Computer Vision (ECCV)},
    year={2024},
}

@Article{kerbl3Dgaussians,
      author       = {Kerbl, Bernhard and Kopanas, Georgios and Leimk{\"u}hler, Thomas and Drettakis, George},
      title        = {3D Gaussian Splatting for Real-Time Radiance Field Rendering},
      journal      = {ACM Transactions on Graphics},
      number       = {4},
      volume       = {42},
      month        = {July},
      year         = {2023},
      url          = {https://repo-sam.inria.fr/fungraph/3d-gaussian-splatting/}
}

@inproceedings{mildenhall2020nerf,
  title={NeRF: Representing Scenes as Neural Radiance Fields for View Synthesis},
  author={Ben Mildenhall and Pratul P. Srinivasan and Matthew Tancik and Jonathan T. Barron and Ravi Ramamoorthi and Ren Ng},
  year={2020},
  booktitle={ECCV},
}

@misc{wu2020visual,
      title={Visual Transformers: Token-based Image Representation and Processing for Computer Vision}, 
      author={Bichen Wu and Chenfeng Xu and Xiaoliang Dai and Alvin Wan and Peizhao Zhang and Zhicheng Yan and Masayoshi Tomizuka and Joseph Gonzalez and Kurt Keutzer and Peter Vajda},
      year={2020},
      eprint={2006.03677},
      archivePrefix={arXiv},
      primaryClass={cs.CV}
}

@misc{theia-manual,
  author = {Chris Sweeney},
  title = {Theia Multiview Geometry Library: Tutorial \& Reference},
  howpublished = "\url{http://theia-sfm.org}",
}

@inproceedings{lindenberger2021pixelperfect,
    author = {Lindenberger, P. and others},
    title = {Pixel-Perfect Structure-from-Motion with Featuremetric Refinement},
    booktitle = {International Conference on Computer Vision (ICCV)},
    year = {2021},
    url = {https://www.cs.jhu.edu/~misha/ReadingSeminar/Papers/Lindenberger21.pdf}
}

@inproceedings{zhou2023continual,
  author = {Zhou, Wenxuan and Zhang, Sheng and Naumann, Tristan and Chen, Muhao and Poon, Hoifung},
  title = {Continual Contrastive Finetuning Improves Low-Resource Relation Extraction},
  booktitle = {Proceedings of the 61st Annual Meeting of the Association for Computational Linguistics (Volume 1: Long Papers)},
  year = {2023},
  address = {Toronto, Canada},
  url = {https://aclanthology.org/2023.acl-long.730},
  publisher = {Association for Computational Linguistics}
}

@inproceedings{choi2016redwood,
  title={A large dataset of object scans},
  author={Choi, Sungjoon and Zhou, Qian-Yi and Koltun, Vladlen},
  booktitle={Proceedings of the IEEE Conference on Computer Vision and Pattern Recognition (CVPR)},
  year={2016},
  pages={4040--4048},
  url={https://redwood-data.org/indoor/dataset.html}
}
\bibliographystyle{acl_natbib}


\end{document}